\documentclass[conference]{IEEEtran}
\IEEEoverridecommandlockouts
\usepackage{cite}
\usepackage{amsmath,amssymb,amsfonts}
\usepackage{algorithmic}
\usepackage{graphicx}
\usepackage{textcomp}
\usepackage{xcolor}

\def\b{\ensuremath\boldsymbol}

\usepackage{url}
\usepackage{balance}

\def\BibTeX{{\rm B\kern-.05em{\sc i\kern-.025em b}\kern-.08em
    T\kern-.1667em\lower.7ex\hbox{E}\kern-.125emX}}
\begin{document}
\bstctlcite{IEEEexample:BSTcontrol}

\title{Fisher Discriminant Triplet and Contrastive Losses \\for Training Siamese Networks}



\author{\IEEEauthorblockN{Benyamin Ghojogh\IEEEauthorrefmark{1},
Milad Sikaroudi\IEEEauthorrefmark{2}, Sobhan Shafiei\IEEEauthorrefmark{2}, \\
H.R. Tizhoosh\IEEEauthorrefmark{2},~\IEEEmembership{Senior Member,~IEEE}, 
Fakhri Karray\IEEEauthorrefmark{1},~\IEEEmembership{Fellow,~IEEE},
Mark Crowley\IEEEauthorrefmark{1}\thanks{Accepted (to appear) in International Joint Conference on Neural Networks (IJCNN) 2020, IEEE, in IEEE World Congress on Computational Intelligence (WCCI) 2020.}
}
\IEEEauthorblockA{\IEEEauthorrefmark{1}Department of Electrical and Computer Engineering, University of Waterloo, Waterloo, ON, Canada \\
\IEEEauthorrefmark{2}Kimia Lab, University of Waterloo, Waterloo, ON, Canada \\
Emails: \{bghojogh, msikaroudi, s7shafie, tizhoosh, karray, mcrowley\}@uwaterloo.ca
}}


\maketitle

\begin{abstract}
Siamese neural network is a very powerful architecture for both feature extraction and metric learning. It usually consists of several networks that share weights. The Siamese concept is topology-agnostic and can use any neural network as its backbone. The two most popular loss functions for training these networks are the triplet and contrastive loss functions. In this paper, we propose two novel loss functions, named Fisher Discriminant Triplet (FDT) and Fisher Discriminant Contrastive (FDC). The former uses anchor-neighbor-distant triplets while the latter utilizes pairs of anchor-neighbor and anchor-distant samples. The FDT and FDC loss functions are designed based on the statistical formulation of the Fisher Discriminant Analysis (FDA), which is a linear subspace learning method. Our experiments on the MNIST and two challenging and publicly available histopathology datasets show the effectiveness of the proposed loss functions. 
\end{abstract}


\section{Introduction}

\begin{figure}[!t]
\centering
\includegraphics[width=3in]{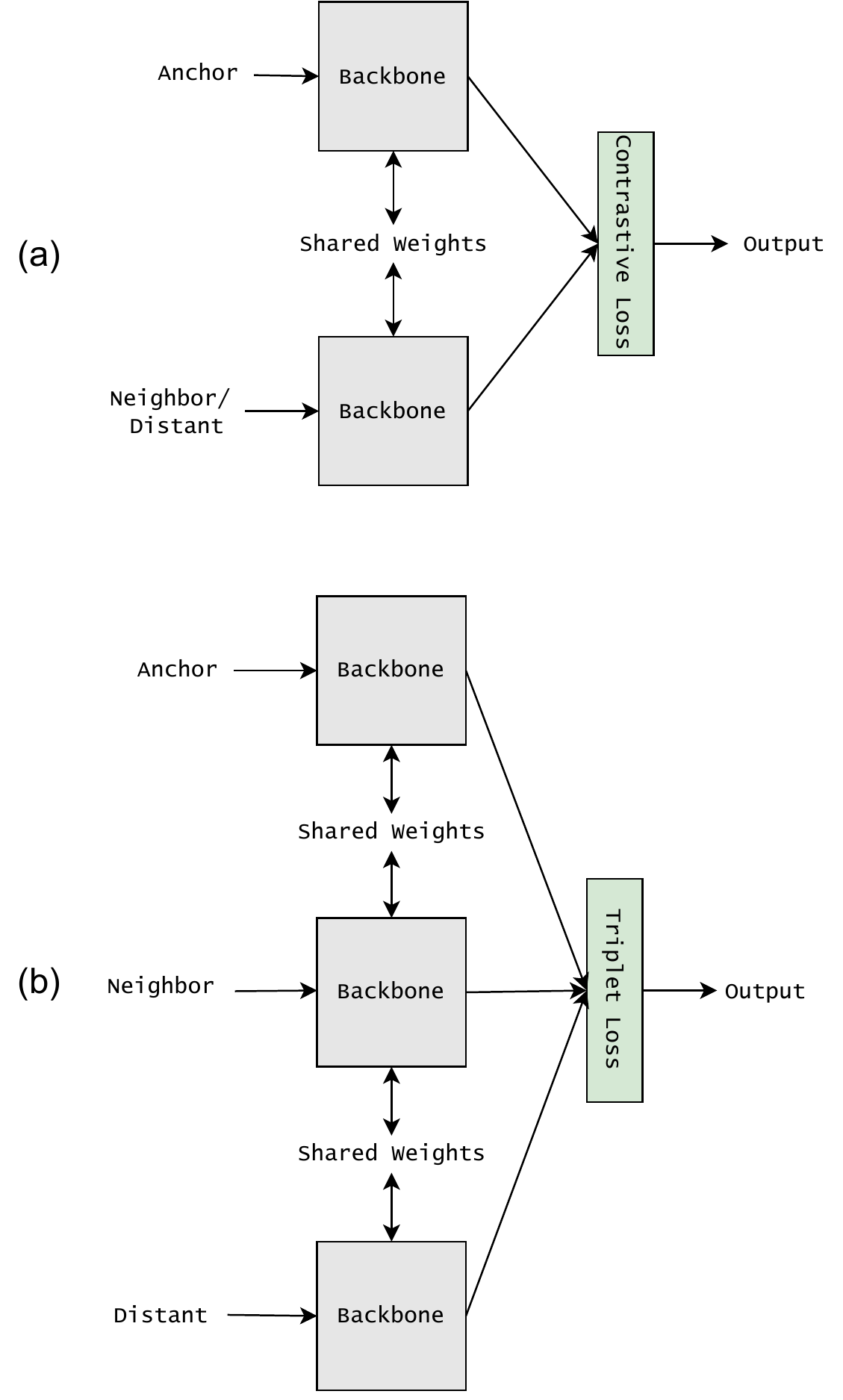}
\caption{Siamese network with (a) contrastive and (b) triplet loss functions.}
\label{figure_Siamese}
\end{figure}

Siamese neural networks have been found very effective for feature extraction \cite{ye2019unsupervised}, metric learning \cite{kumar2016learning}, few-shot learning \cite{koch2015siamese}, and feature tracking \cite{he2018twofold}. 
A Siamese network includes several, typically two or three, backbone neural networks which share weights \cite{schroff2015facenet} (see Fig. \ref{figure_Siamese}). 
Different loss functions have been proposed for training a Siamese network. Two commonly used ones are triplet loss \cite{schroff2015facenet} and contrastive loss \cite{hadsell2006dimensionality} which are displayed in Fig. \ref{figure_Siamese}. 

We generally start with considering two samples named an anchor and one of its neighbors from the same class, and two more samples named the same anchor and a distant counterpart from a different class. The triplet loss considers the anchor-neighbor-distant triplets while the contrastive loss deals with the anchor-neighbor and anchor-distant pairs of samples. The main idea of these loss functions is to pull the samples of every class toward one another and push the samples of different classes away from each other in order to improve the classification results and hence the generalization capability. We will introduce these losses in Section \ref{section_background_Siamese} in more depth. 

The Fisher Discriminant Analysis (FDA) \cite{friedman2001elements} was first proposed in \cite{fisher1936use}. FDA is a linear method based on generalized eigenvalue problem \cite{ghojogh2019roweis} and tries to find an embedding subspace that decreases the variance of each class while increases the variance between the classes. As can be observed, there is a similar intuition behind the concepts of both the Siamese network and the FDA, where they try to embed the data in a way that the samples of each class collapse close together \cite{globerson2006metric} but the classes fall far away. 
Although FDA is a well-known statistical method, it has been recently noticed in the literature of deep learning\cite{diaz2017deep,diaz2019deep}. 

Noticing the similar intuition behind the Siamese network and FDA, we propose two novel loss functions for training Siamese networks, which are inspired by the theory of the FDA. We consider the intra- and inter-class scatters of the triplets instead of their $\ell_2$ norm distances. The two proposed loss functions are Fisher Discriminant Triplet (FDT) and Fisher Discriminant Contrastive (FDC) losses, which correspond to triplet and contrastive losses, respectively. Our experiments show that these loss functions exhibit very promising behavior for training Siamese networks. 

The remainder of the paper is organized as follows: Section \ref{section_background} reviews the foundation of the Fisher criterion, FDA, Siamese network, triplet loss, and contrastive loss. In Sections \ref{section_FDT} and \ref{section_FDC}, we propose the FDT and FDC loss functions for training Siamese networks, respectively. In Section \ref{section_experiments}, we report multiple experiments on different benchmark datasets to demonstrate the effectiveness of the proposed losses. Finally, Section \ref{section_conclusion} concludes the paper. 

\section{Background}\label{section_background}

\subsection{Fisher Criterion}
Assume the data include $c$ classes where the $k$-th class, with the sample size $n_k$, is denoted by $\{\b{x}_i^{(k)}\}_{i=1}^{n_k}$. Let the dimensionality of data be $d$.
Consider a $p$-dimensional subspace (with $p \leq d$) onto which the data are projected. We can define intra- (within) and inter-class (between) scatters as the scatter of projected data in and between the classes. 
The Fisher criterion is increasing and decreasing with the intra- and inter-class scatters, respectively; hence, by maximizing it, one can aim to maximize the inter-class scatter of projected data while minimizing the intra-class scatter. 
There exist different versions of the Fisher criterion \cite{fukunaga2013introduction}. 
Suppose $\b{U} \in \mathbb{R}^{d \times p}$ is the projection matrix onto the subspace, then the trace of the matrix $\b{U}^\top \b{S}\, \b{U}$ can be interpreted as the variance of the projected data \cite{ghojogh2019fisher}. Based on this interpretation, the most popular Fisher criterion is defined as follows \cite{fisher1936use,xu2006analysis}
\begin{align}\label{equation_Fisher_criterion_1}
J := \frac{\textbf{tr}(\b{U}^\top \b{S}_B\, \b{U})}{\textbf{tr}(\b{U}^\top \b{S}_W\, \b{U})},
\end{align}
where $\textbf{tr}(\cdot)$ denotes the trace of matrix and $\b{S}_B \in \mathbb{R}^{d \times d}$ and $\b{S}_W \in \mathbb{R}^{d \times d}$ are the inter- and intra-class scatter matrices, respectively, defined as
\begin{align}
\b{S}_W &:= \sum_{k=1}^c \sum_{i=1}^{n_k}  \sum_{j=1}^{n_k} (\b{x}_i^{(k)} - \b{x}_j^{(k)}) (\b{x}_i^{(k)} - \b{x}_j^{(k)})^\top, \textrm{and}\\
\b{S}_B &:= \sum_{k=1}^c \sum_{\ell=1, \ell \neq k}^c \sum_{i=1}^{n_k}  \sum_{j=1}^{n_\ell} (\b{x}_i^{(k)} - \b{x}_j^{(\ell)}) (\b{x}_i^{(k)} - \b{x}_j^{(\ell)})^\top.
\end{align}
Some other versions of Fisher criterion are \cite{fukunaga2013introduction}
\begin{align}
& J := \b{S}_W^{-1} \b{S}_B,  \textrm{ and} \label{equation_Fisher_criterion_2} \\
& J := \textbf{tr}(\b{U}^\top \b{S}_B\, \b{U}) - \textbf{tr}(\b{U}^\top \b{S}_W\, \b{U}),  \label{equation_Fisher_criterion_3}
\end{align}
where the former is because the solution to maximizing \eqref{equation_Fisher_criterion_1} is the generalized eigenvalue problem $(\b{S}_B, \b{S}_W)$ (see Section \ref{section_FDA}) whose solution can be the eigenvectors of $\b{S}_W^{-1} \b{S}_B$ \cite{ghojogh2019eigenvalue}. The reason for latter is because \eqref{equation_Fisher_criterion_1} is a Rayleigh-Ritz quotient \cite{parlett1998symmetric} and its denominator can be set to a constant \cite{ghojogh2019fisher}. The Lagrange relaxation of the optimization would be similar to \eqref{equation_Fisher_criterion_3}.  

\subsection{Fisher Discriminant Analysis}\label{section_FDA}

FDA \cite{fisher1936use,friedman2001elements} is defined as a linear transformation which maximizes the criterion function (\ref{equation_Fisher_criterion_1}). This criterion is a generalized Rayleigh-Ritz quotient \cite{parlett1998symmetric} and we may recast the problem to \cite{ghojogh2019fisher}
\begin{equation}\label{equation_optimization_FDA}
\begin{aligned}
& \underset{\b{U}}{\text{maximize}}
& & \textbf{tr}(\b{U}^\top \b{S}_B\, \b{U}), \\
& \text{subject to}
& & \b{U}^\top \b{S}_W\, \b{U} = \b{I},
\end{aligned}
\end{equation}
where $\b{I}$ is the identity matrix.
The Lagrange relaxation of the problem can be written as follows
\begin{align}
\mathcal{L} = \textbf{tr}(\b{U}^\top \b{S}_B\, \b{U}) - \textbf{tr}\big(\b{\Lambda}^\top (\b{U}^\top \b{S}_W\, \b{U} - \b{I})\big),
\end{align}
where $\b{\Lambda}$ is a diagonal matrix which includes the Lagrange multipliers \cite{boyd2004convex}. 
Setting the derivative of Lagrangian to zero gives
\begin{align}
& \frac{\partial \mathcal{L}}{\partial \b{U}} = 2\b{S}_B \b{U} - 2\b{S}_W\b{U} \b{\Lambda} \overset{\text{set}}{=} \b{0}\!\implies\!\b{S}_B\, \b{U} = \b{S}_W\, \b{U} \b{\Lambda}, \label{equation_scatter_generalized_eigendecomposition}
\end{align}
which is the generalized eigenvalue problem $(\b{S}_B, \b{S}_W)$ where the columns of $\b{U}$ and the diagonal of $\b{\Lambda}$ are the eigenvectors and eigenvalues, respectively \cite{ghojogh2019eigenvalue}. 
The column space of $\b{U}$ is the FDA subspace. 

\subsection{Siamese Network and Loss Functions}\label{section_background_Siamese}

\subsubsection{Siamese Network}

Siamese network is a set of several (typically two or three) networks which share weights with each other \cite{schroff2015facenet} (see Fig. \ref{figure_Siamese}). The weights are trained using a loss based on anchor, neighbor (positive), and distant (negative) samples, where anchor and neighbor belong to the same class, but the anchor and distant tiles are in different classes. We denote the anchor, neighbor, and distant samples by $\b{x}_a$, $\b{x}_n$, and $\b{x}_d$, respectively. The loss functions used to train a Siamese network usually make use of the anchor, neighbor, and distant samples, trying to pull the anchor and neighbor towards one another and simultaneously push the anchor and distant tiles away from each other. In the following, two different loss functions are introduced for training Siamese networks. 

\subsubsection{Triplet Loss}

The triplet loss uses anchor, neighbor, and distant. Let $\mathbf{f}(\b{x})$ be the output (i.e., embedding) of the network for the input $\b{x}$. The triplet loss tries to reduce the distance of anchor and neighbor embeddings and desires to increase the distance of anchor and distant embeddings. As long as the distances of anchor-distant pairs get larger than the distances of anchor-neighbor pairs by a margin $\alpha \geq 0$, the desired embedding is obtained. The triplet loss, to be minimized, is defined as \cite{schroff2015facenet}
\begin{equation}
\ell_\text{t}\!=\!\sum_{i=1}^b\! \Big[\lVert\mathbf{f}(\b{x}_a^i)\!-\!\mathbf{f}(\b{x}_n^i)\rVert_2^2\!-\!\lVert\mathbf{f}(\mathbf{x}_a^i)\!-\!\mathbf{f}(\b{x}_d^i)\rVert_2^2\!+\!\alpha\Big]_{+}
\end{equation}
where $\b{x}^i$ is the $i$-th triplet sample in the mini-batch, $b$ is the mini-batch size, $[z]_+ := \max(z,0)$, and $||\cdot||_2$ denotes the $\ell_2$ norm.

\subsubsection{Contrastive Loss}

The contrastive loss uses pairs of samples which can be anchor and neighbor or anchor and distant. If the samples are anchor and neighbor, they are pulled towards each other; otherwise, their distance is increased. 
In other words, the contrastive loss performs like the triplet loss but one by one rather than simultaneously. 
The desired embedding is obtained when the anchor-distant distances get larger than the anchor-neighbor distances by a margin of $\alpha$. 
This loss, to be minimized, is defined as \cite{hadsell2006dimensionality}
\begin{equation}
\label{equation_contrastive}
\ell_\text{c}\!=\!\sum_{i=1}^b \Big[ (1\!-\!y) ||\mathbf{f}(\b{x}_1^i) - \mathbf{f}(\b{x}_2^i)||_2^2 + y\, \big[\! -\!||\mathbf{f}(\b{x}_1^i) - \mathbf{f}(\b{x}_2^i)||_2^2 + \alpha \big]_+ \Big]
\end{equation}
where $y$ is zero and one when the pair $\{\b{x}_1^i, \b{x}_2^i\}$ is anchor-neighbor and anchor-distant, respectively. 

\section{The Proposed Loss Functions}
\label{section_proposed_losses}

\subsection{Network Structure}
\label{section_network_structure}

\begin{figure}[!t]
\centering
\includegraphics[width=3.2in]{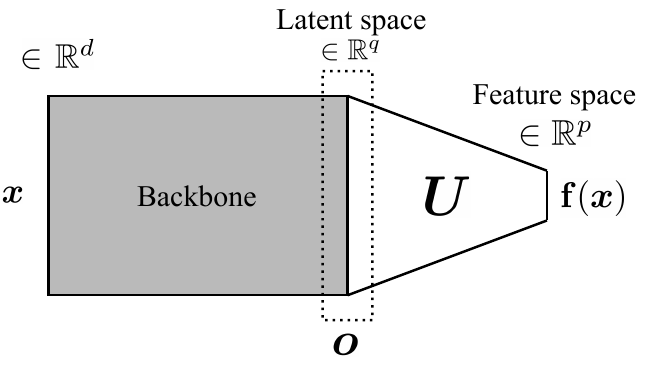}
\caption{The network structure for the proposed loss functions.}
\label{figure_network_structure}
\end{figure}

Consider any arbitrary neural network as the backbone. This network can be either a multi-layer perception or a convolutional network. Let $q$ be the number of its output neurons, i.e., the dimensionality of its embedding space. We add a fully connected layer after the $q$-neurons layer to a new embedding space (output layer) with $p \leq q$ neurons. Denote the weights of this layer by $\b{U} \in \mathbb{R}^{q \times p}$. We name the first $q$-dimensional embedding space as the \textit{latent space} and the second $p$-dimensional embedding space as the \textit{feature space}. 
Our proposed loss functions are network-agnostic as they can be used for any network structure and topology of the backbone. The overall network structure for the usage of the proposed loss functions is depicted in Fig. \ref{figure_network_structure}.

Consider a triplet $\{\b{x}_a,\b{x}_n,\b{x}_d\in\mathbb{R}^d\}$ or a pair $\{\b{x}_1,\b{x}_2\in\mathbb{R}^d\}$. We feed the triplet or pair to the network. We denote the latent embedding of data by $\{\b{o}_a, \b{o}_n, \b{o}_d \in \mathbb{R}^q\}$ and $\{\b{o}_1, \b{o}_2 \in \mathbb{R}^q\}$ while the feature embedding of data is denoted by $\{\mathbf{f}(\b{x}_a), \mathbf{f}(\b{x}_n), \mathbf{f}(\b{x}_d) \in \mathbb{R}^p\}$ and $\{\mathbf{f}(\b{x}_1), \mathbf{f}(\b{x}_2) \in \mathbb{R}^p\}$. The last layer of network is projecting the latent embedding to the feature space where the activation function of the last layer is linear because of unsupervised feature extraction. Hence, the last layer acts as a linear projection $\mathbf{f}(\b{x}) = \b{U}^\top \b{o}$. 

 During the training, the latent space is adjusted to extract some features; however, the last-layer projection fine-tunes the latent features in order to have better discriminative features. In Section \ref{section_experiments_latent_features}, we show the results of experiments to demonstrate this. 

\subsection{Fisher Discriminant Triplet Loss}\label{section_FDT}

As in neural networks the loss function is usually minimized, we minimize the negative of Fisher criterion where we use \eqref{equation_Fisher_criterion_3} as the criterion:
\begin{align}
&\underset{\b{U}}{\text{minimize}} ~~~ -J = \textbf{tr}(\b{U}^\top \b{S}_W\, \b{U}) - \textbf{tr}(\b{U}^\top \b{S}_B\, \b{U}).
\end{align}
This problem is ill-defined because by increasing the total scatter of embedded data, the inter-class scatter also gets larger and this objective function gets decreased. Therefore, the embedding space scales up and explodes gradually to increase the term $\textbf{tr}(\b{U}^\top \b{S}_B\, \b{U})$.
In order to control this issue, we penalize the total scatter of the embedded data, denoted by $\b{S}_T \in \mathbb{R}^{d \times d}$:
\begin{align}
&\underset{\b{U}}{\text{min}} ~~ \textbf{tr}(\b{U}^\top\!\b{S}_W\, \b{U})\!-\! \textbf{tr}(\b{U}^\top \b{S}_B\, \b{U})\!+\! \epsilon\, \textbf{tr}(\b{U}^\top \b{S}_T\, \b{U}),
\end{align}
where $\epsilon \in (0,1)$ is the regularization parameter. The total scatter can be considered as the summation of the inter- and intra-class scatters \cite{ye2007least}:
\begin{align}\label{equation_S_T_as_sum_of_scatters}
\b{S}_T := \b{S}_B + \b{S}_W.
\end{align}
Hence, we have:
\begin{align}
\textbf{tr}(\b{U}^\top &\b{S}_W\, \b{U}) - \textbf{tr}(\b{U}^\top \b{S}_B\, \b{U}) + \epsilon\, \textbf{tr}(\b{U}^\top \b{S}_T\, \b{U}) \nonumber \\
&= \textbf{tr}\big(\b{U}^\top (\b{S}_W - \b{S}_B + \epsilon\, \b{S}_T)\, \b{U}\big) \nonumber \\
&\overset{(\ref{equation_S_T_as_sum_of_scatters})}{=} \textbf{tr}\big(\b{U}^\top ((\epsilon+1)\, \b{S}_W + (\epsilon-1)\, \b{S}_B)\, \b{U}\big) \nonumber \\
&\overset{(a)}{=} (2 - \lambda)\,\textbf{tr}(\b{U}^\top \b{S}_W\, \b{U}) - \lambda\, \textbf{tr}(\b{U}^\top \b{S}_B\, \b{U}),
\end{align}
where $(a)$ is because $(0,1) \ni \lambda := 1 - \epsilon$. 
It is recommended for $\epsilon$ and $\lambda$ to be close to one and zero, respectively because the total scatter should be controlled not to explode. For example, a good value can be $\lambda = 0.1$. 

We want the inter-class scatter term to get larger than the intra-class scatter term by a margin $\alpha > 0$. Hence, the FDT loss, to be minimized, is defined as:
\begin{equation}
\ell_\text{fdt}\!=\!\Big[ (2-\lambda)\, \textbf{tr}(\b{U}^\top \b{S}_W\, \b{U})\!-\!\lambda\, \textbf{tr}(\b{U}^\top \b{S}_B\, \b{U})\!+\!\alpha\Big]_+
\end{equation}
where we defer the mathematical definition of intra- and inter-class scatter matrices in our loss functions to Section \ref{section_scatter_matrices_in_FDT}. 

\subsection{Fisher Discriminant Contrastive Loss}\label{section_FDC}

Rather than the triplets of data, we can consider the pairs of samples. For this goal, we propose the FDC loss function defined as
\begin{equation}
\ell_\text{fdc}\!=\!(2-\lambda)\, \textbf{tr}(\b{U}^\top \widetilde{\b{S}}_W\, \b{U}) + \big[\! -\!\lambda\, \textbf{tr}(\b{U}^\top \widetilde{\b{S}}_B\, \b{U}) + \alpha \big]_+
\end{equation}
where the intra- and inter-class scatter matrices, which will be defined in Section \ref{section_scatter_matrices_in_FDT}, consider the anchor-neighbor and anchor-distant pairs. 

\subsection{Intra- and Inter-Class Scatters}
\label{section_scatter_matrices_in_FDT}

\subsubsection{Scatter Matrices in FDT}

Let the output embedding of the backbone, i.e. the second-to-last layer of total structure, be denoted by $\b{o} \in \mathbb{R}^q$. We call this embedding the \textit{latent embedding}. Consider the latent embeddings of anchor, neighbor, and distant, denoted by $\b{o}_a$, $\b{o}_n$, and $\b{o}_d$, respectively. If we have a mini-batch of $b$ triplets, we can define $\mathbb{R}^{q \times b} \ni \b{O}_W := [\b{o}_a^1 - \b{o}_n^1, \dots, \b{o}_a^b - \b{o}_n^b]$ and $\mathbb{R}^{q \times b} \ni \b{O}_B := [\b{o}_a^1 - \b{o}_d^b, \dots, \b{o}_a^b - \b{o}_d^b]$ where $\b{o}^i$ is the $i$-th sample in the mini-batch. 
The intra- and inter-class scatter matrices are, respectively, defined as
\begin{align}
\mathbb{R}^{q \times q} \ni \b{S}_W &:= \sum_{i=1}^b (\b{o}_a^{i} - \b{o}_n^{i}) (\b{o}_a^{i} - \b{o}_n^{i})^\top = \b{O}_W\, \b{O}_W^\top, \\
\mathbb{R}^{q \times q} \ni \b{S}_B &:= \sum_{i=1}^b (\b{o}_a^{i} - \b{o}_d^{i}) (\b{o}_a^{i} - \b{o}_d^{i})^\top = \b{O}_B\, \b{O}_B^\top.
\end{align}
The ranks of the intra- and inter-class scatters are $\min(q, b-1)$. As the subspace of FDA can be interpreted as the eigenspace of $\b{S}_W^{-1}\b{S}_B$, the rank of the subspace would be $\min(q, b-1) = b-1$ because we usually have $b < q$. In order to improve the rank of the embedding subspace, we slightly strengthen the main diagonal of the scatter matrices \cite{mika1999fisher}
\begin{align}
&\b{S}_W := \b{O}_W\, \b{O}_W^\top + \mu_W\, \b{I}, \\
&\b{S}_B := \b{O}_B\, \b{O}_B^\top + \mu_B\, \b{I},
\end{align}
where $\mu_W, \mu_B >0$ are small positive numbers, e.g., $10^{-4}$.
Hence, the embedding subspace becomes full rank with $q \geq p$.

\subsubsection{Scatter Matrices in FDC}
As in the regular contrastive loss, we consider the pairs of anchor-neighbor and anchor-distant for the FDC loss. 
Let $y$ be zero and one when the pair $\{\b{x}_1^i, \b{x}_2^i\}$ is an anchor-neighbor or anchor-distant pair, respectively. The latent embedding of this pair is denoted by $\{\b{o}_1^i, \b{o}_2^i\}$. The intra- and inter-class scatter matrices in the FDC loss are, respectively, defined as
\begin{align}
\widetilde{\b{S}}_W &:= \sum_{i=1}^b (1-y) (\b{o}_1^{i} - \b{o}_2^{i}) (\b{o}_1^{i} - \b{o}_2^{i})^\top + \mu_W\, \b{I} \nonumber \\
&= \widetilde{\b{O}}_W\, \widetilde{\b{O}}_W^\top + \mu_W\, \b{I}, \\
\widetilde{\b{S}}_B &:= \sum_{i=1}^b y (\b{o}_1^{i} - \b{o}_2^{i}) (\b{o}_1^{i} - \b{o}_2^{i})^\top + \mu_B\, \b{I} \nonumber \\
&= \widetilde{\b{O}}_B\, \widetilde{\b{O}}_B^\top + \mu_B\, \b{I},
\end{align}
where $\widetilde{\b{O}}_W\!:=\![\{\b{o}_1^i - \b{o}_2^i \,|\, y=0\}]$ and $\widetilde{\b{O}}_B\!:=\![\{\b{o}_1^i - \b{o}_2^i \,|\, y\!=\!1\}]$. 

Note that in both FDT and FDC loss functions, there exist the weight matrix $\b{U}$ and the intra- and inter-class scatter matrices. By back-propagation, both the last layer and the previous layers are trained because $\b{U}$ in loss affects the last layer, and the scatter matrices in loss impact all the layers. 

\section{Experiments}
\label{section_experiments}

\subsection{Datasets}

For the experiments, we used three public datasets, i.e., MNIST and two challenging histopathology datasets. In the following, we introduce these datasets. The MNIST images have one channel, but the histopathology images exhibit color in three channels. 

\textbf{MNIST dataset --} The MNIST dataset \cite{web_mnist_dataset} includes 60,000 training images and 10,000 test images of size $28 \times 28$ pixels. We created a dataset of 500 triplets from the training data to test the proposed loss functions for a small training sample size. 

\begin{figure*}[!t]
\centering
\includegraphics[width=7in]{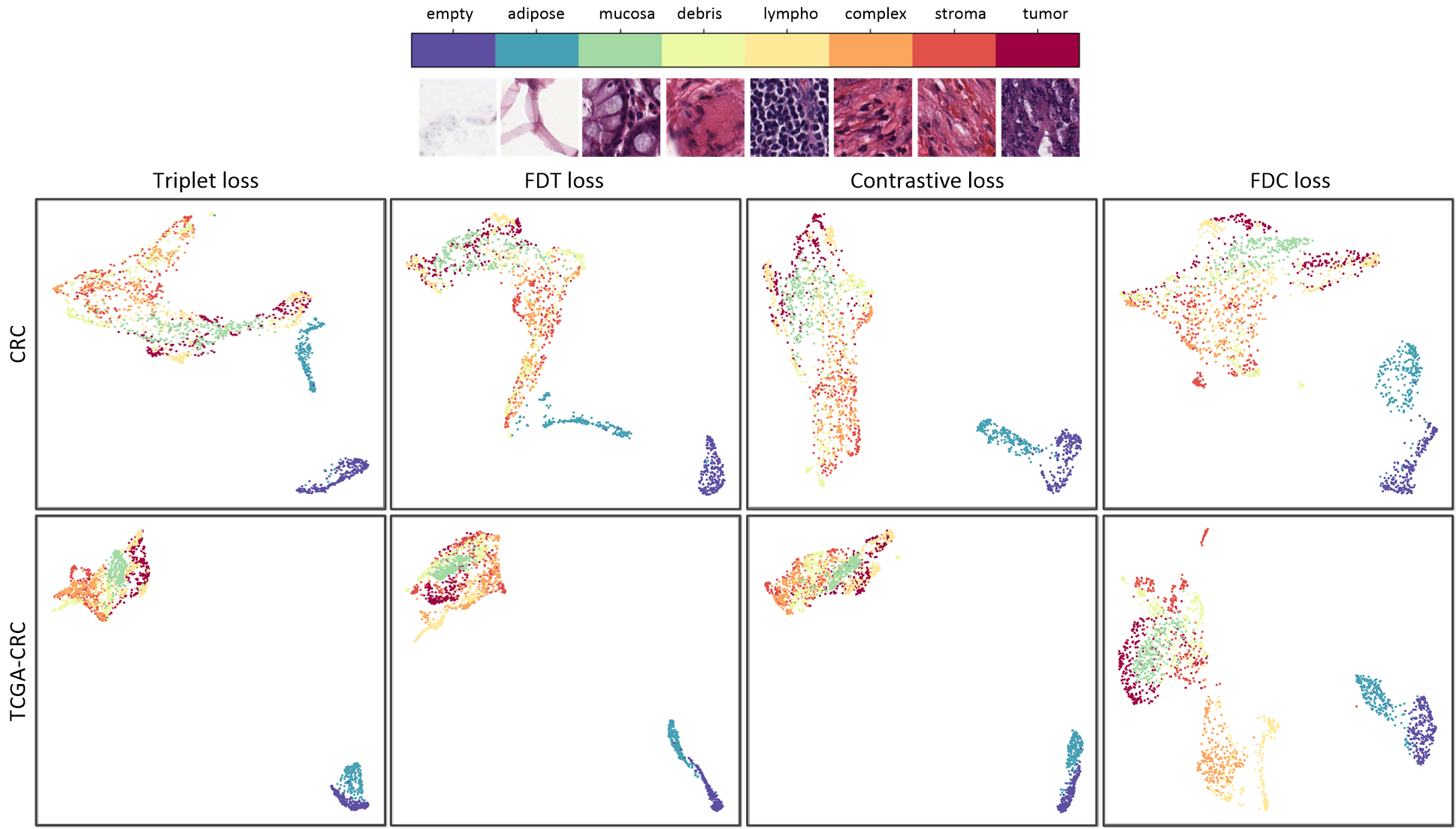}
\caption{Embedding of the CRC test data for different loss functions (top row: CRC, bottom row: TCGA).}
\label{figure_embeddings_CRC}
\end{figure*}

\textbf{CRC dataset --} The first histopathology dataset is the Colorectal Cancer (CRC) dataset \cite{kather2016multi}. It contains tissue patches from different tissue types of colorectal cancer tissue slides. The tissue types are background (empty), adipose tissue, mucosal glands, debris, immune cells (lymphoma), complex stroma, simple stroma, and tumor epithelium. Some sample patches of CRC tissue types can be seen in Fig. \ref{figure_embeddings_CRC}.
We split data into train/test sets with 60\%--40\% portions.  Using the training set, we extracted 22,528 triplets by considering the tissue types as the classes. 

\textbf{TCGA dataset --} The second histopathology dataset is The Cancer Genome Atlas (TCGA) dataset \cite{cooper2018pancancer}. TCGA Whole Slide Images (WSIs) come from 25 different organs for 32 different cancer subtypes. We use the three most common sites, which are prostate, gastrointestinal, and lung \cite{cooper2018pancancer,kalra2020pan}. These organs have a total of 9 cancer subtypes, i.e., Prostate adenocarcinoma (PRAD), Testicular germ cell tumors (TGCT), Oesophageal carcinoma (ESCA), Stomach adenocarcinoma (STAD), Colonic adenocarcinoma (COAD), Rectal adenocarcinoma (READ), Lung adenocarcinoma (LUAD), Lung squamous cell carcinoma (LUSC), and Mesothelioma (MESO). 
By sampling patches from slides, we extracted 22,528 triplets to test the proposed losses with a large triplet sample size. The anchor and neighbor patches were selected from one WSI, but we used four ways of extraction of the distant patch, i.e., from the same WSI but far from the anchor, from another WSI of the same cancer subtype as an anchor, from another cancer subtype but the same anatomic site as anchor, and from another anatomic site.

\begin{figure*}[!t]
\centering
\includegraphics[width=7in]{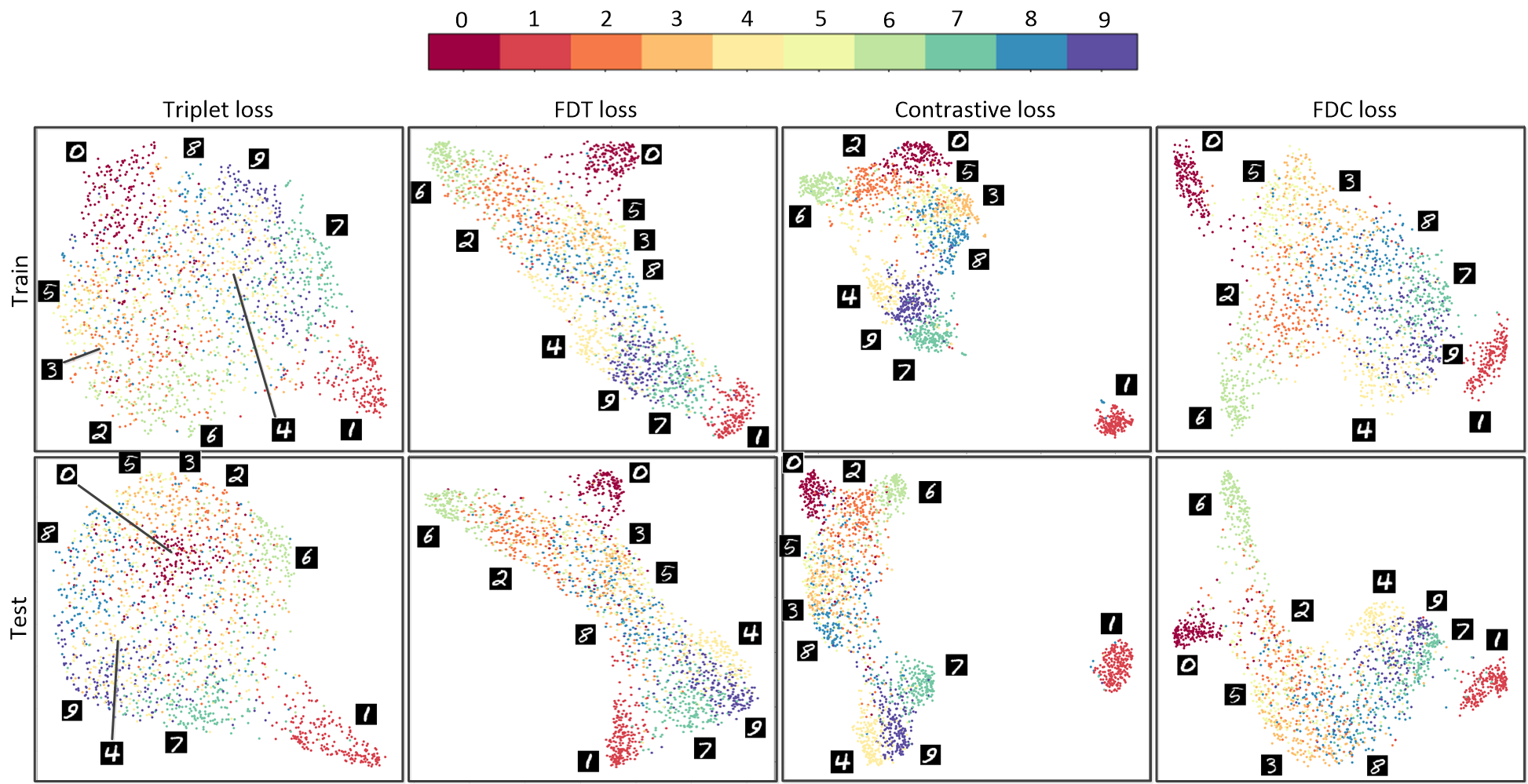}
\caption{Embedding of the training and test sets of MNIST dataset in the feature spaces of different loss functions.}
\label{figure_embeddings_mnist}
\end{figure*}

\subsection{Visual Comparison of Emebddings}

In our experiments, we used ResNet-18 \cite{he2016deep} as the backbone in our Siamese network structure (see Fig. \ref{figure_network_structure}). 
In our experiments, we set $q=300$, $p=128$, $b=32$, and $\alpha=0.25$. The learning rate was set to $10^{-5}$ in all experiments.

\textbf{Embedding of MNIST data --} The embeddings of the train/test sets of the MNIST dataset in the feature spaces of different loss functions are illustrated in Fig. \ref{figure_embeddings_mnist} where $\lambda=0.1$ was used for FDT and FDC. 
We used the Uniform Manifold Approximation and Projection (UMAP) \cite{mcinnes2018umap} for visualizing the 128-dimensional embedded data. As can be seen, both embeddings of train and test data by the FDT loss are much more discriminating than the embedding of triplet loss. On the other hand, comparing the embeddings of contrastive and FDC losses shows that their performances are both good enough as the classes are well separated.  
 Interestingly, the similar digits usually are embedded as close classes in the feature space, and this shows the meaningfulness of the trained subspace. For example, the digit pairs (3, 8), (1, 7), and (4, 9) with the second writing format of digit four can transition into each other by slight changes, and that is why they are embedded close together. 

\textbf{Embedding of histopathology data --} For embedding of the histopathology data, we performed two different experiments. 
In the first experiment, we trained and tested the Siamese network using the CRC data. The second experiment was to train the Siamese network using TCGA data and test it using the CRC test set. The latter, which we denote by TCGA-CRC, is more difficult because it tests generalization of the feature space, which is trained by different data from the test data, although with a similar texture. Figure \ref{figure_embeddings_CRC} shows the embeddings of the CRC test sets in the feature spaces trained by CRC and TCGA data.
The embeddings by all losses, including FDT and FDC, are acceptable, noticing that the histopathology data are hard to discriminate even by a human (see the sample patches in Fig. \ref{figure_embeddings_CRC}). As expected, the empty and adipose data, which are similar, are embedded closely. Comparing the TCGA-CRC embeddings of contrastive and FDC losses shows FDC has discriminated classes slightly better. Overall, the good embedding in TCGA-CRC shows that the proposed losses can train a generalizing feature space, which is very important in histopathology analysis because of the lack of labeled data \cite{jimenez2017analysis}.

\subsection{Numerical Comparison of Embeddings}

In addition to visualization, we can assess the embeddings numerically. 
For the evaluation of the embedded subspaces, we used the 1-Nearest Neighbor (1-NN) search because it is useful to evaluate the subspace by the closeness of the projected data samples. 
The accuracy rates of the 1-NN search for the embedding test data by different loss functions are reported in Table \ref{table_1NN_search}. 
We report the results for different values of $\lambda \in \{0.01, 0.1, 0.8\}$ in order to analyze the effect of this hyper-parameter. 
As the results show, in most cases, the FDT and FDC losses have outperformed the triplet and contrastive losses, respectively. Moreover, we see that $\lambda=0.1$ is often better performing. This can be because the large value of $\lambda$ (e.g., $0.8$) imposes less penalty on the total scatter, which may cause the embedding space to expand gradually. The very small value of $\lambda$ (e.g., $0.01$), on the other hand, puts too much emphasis on the total scatter where the classes do not tend to separate well enough, so they do not increase the total scatter.   

\begin{table}[!t]
\caption{Accuracy of $1$-NN search for different loss functions.}
\label{table_1NN_search}
\centering
\scalebox{1}{    
\begin{tabular}{l || c | c | c }
& MNIST & CRC & TCGA-CRC   \\
\hline
\hline
triplet & 82.21\% & 95.75\% & 95.50\% \\
FDT ($\lambda=0.01$) & 82.76\% & 96.45\% & 97.60\% \\
FDT ($\lambda=0.1$) & 85.74\% & 96.05\% & 96.40\% \\
FDT ($\lambda=0.8$) & 79.59\% & 95.35\% & 95.95\% \\
\hline
contrastive &  89.99\% & 95.55\% & 96.55\% \\
FDC ($\lambda=0.01$) & 78.47\% & 94.25\% & 96.55\%  \\
FDC ($\lambda=0.1$) & 89.00\% & 96.40\% & 98.10\% \\
FDC ($\lambda=0.8$) & 87.71\% & 97.00\% & 97.05\% \\
\hline
\hline
\end{tabular}%
}
\end{table}


\subsection{Comparison of the Latent and Feature Spaces}\label{section_experiments_latent_features}

\begin{figure}[!t]
\centering
\includegraphics[width=1.9in]{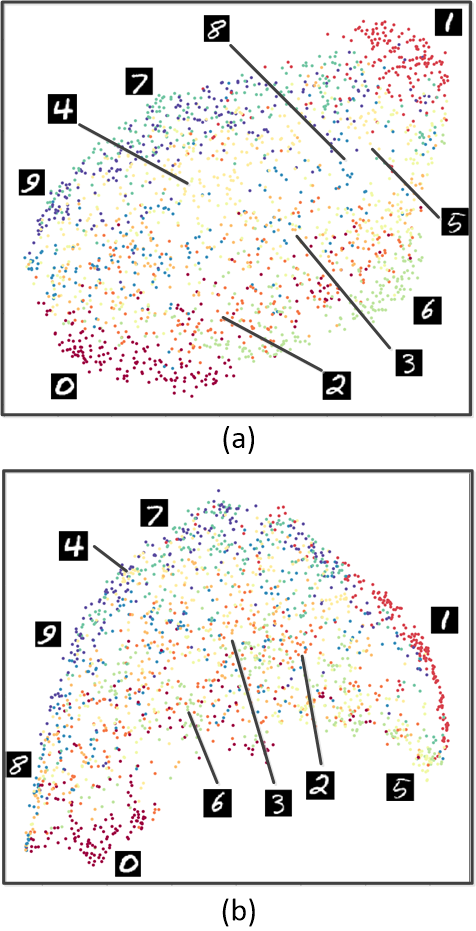}
\caption{The latent embedding of MNIST training for (a) FDT and (b) FDC.}
\label{figure_embeddings_mnist_secondToLast}
\end{figure}

As explained in Section \ref{section_network_structure}, the last layer behaves as a linear projection of the latent space onto the feature space. This projection fine-tunes the embeddings for better discrimination of classes. Figure \ref{figure_embeddings_mnist_secondToLast} shows the latent embedding of the MNIST train set for both FDT and FDC loss functions. Comparing them to the feature embeddings of the MNIST train set in Fig. \ref{figure_embeddings_mnist} shows that the feature embedding discriminates the classes much better than the latent embedding.

\section{Conclusions}
\label{section_conclusion}

In this paper, we proposed two novel loss functions for training Siamese networks. These losses were based on the theory of the FDA, which attempts to decrease the intra-class scatter but increase the inter-class scatter of the projected data. The FDT and FDC losses make use of triplets and pairs of samples, respectively. By experimenting on MNIST and two histopathology datasets, we showed that the proposed losses mostly perform better than the well-known triplet and contrastive loss functions for Siamese networks.


\bibliographystyle{IEEEtran}
\balance
\bibliography{references}

\end{document}